\documentclass[10pt,twocolumn,letterpaper]{article}

\usepackage{cvpr}              
\definecolor{cvprblue}{rgb}{0.21,0.49,0.74}
\usepackage[pagebackref,breaklinks,colorlinks,allcolors=cvprblue]{hyperref}

\def\paperID{6092} 
\def\confName{CVPR}
\def\confYear{2026}

\title{SpriteHand: Real-Time Versatile Hand-Object Interaction with Autoregressive Video Generation}

\author{Zisu Li\\
HKUST\\
Hong Kong SAR\\
{\tt\small zlihe@connect.ust.hk}
\and
Hengye Lyu\\
HKUST (Guangzhou)\\
Guangzhou, China\\
{\tt\small hengyelyu@hkust-gz.edu.cn}
\and
Jiaxin Shi\\
XMax.AI Ltd.\\
Beijing, China\\
{\tt\small jiaxin@xmax.ai}
\and
Yufeng Zeng\\
HKUST (Guangzhou)\\
First line of institution2 address\\
{\tt\small yzeng683@connect.hkust-gz.edu.cn}
\and
Mingming Fan\\
HKUST (Guangzhou)\\
Guangzhou, China\\
{\tt\small mingmingfan@hkust-gz.edu.cn}
\and
Hanwang Zhang\\
Nanyang Technological University\\
Singapore\\
{\tt\small  hanwangzhang@ntu.edu.sg}
\and
Chen Liang\\
HKUST (Guangzhou) \& XMax.AI Ltd. \\
Guangzhou\& Beijing, China\\
{\tt\small  chenliang2@hkust-gz.edu.cn}
}

\begin{document}

\maketitle
\begin{figure}[t]
  \centering
  \includegraphics[width=\columnwidth]{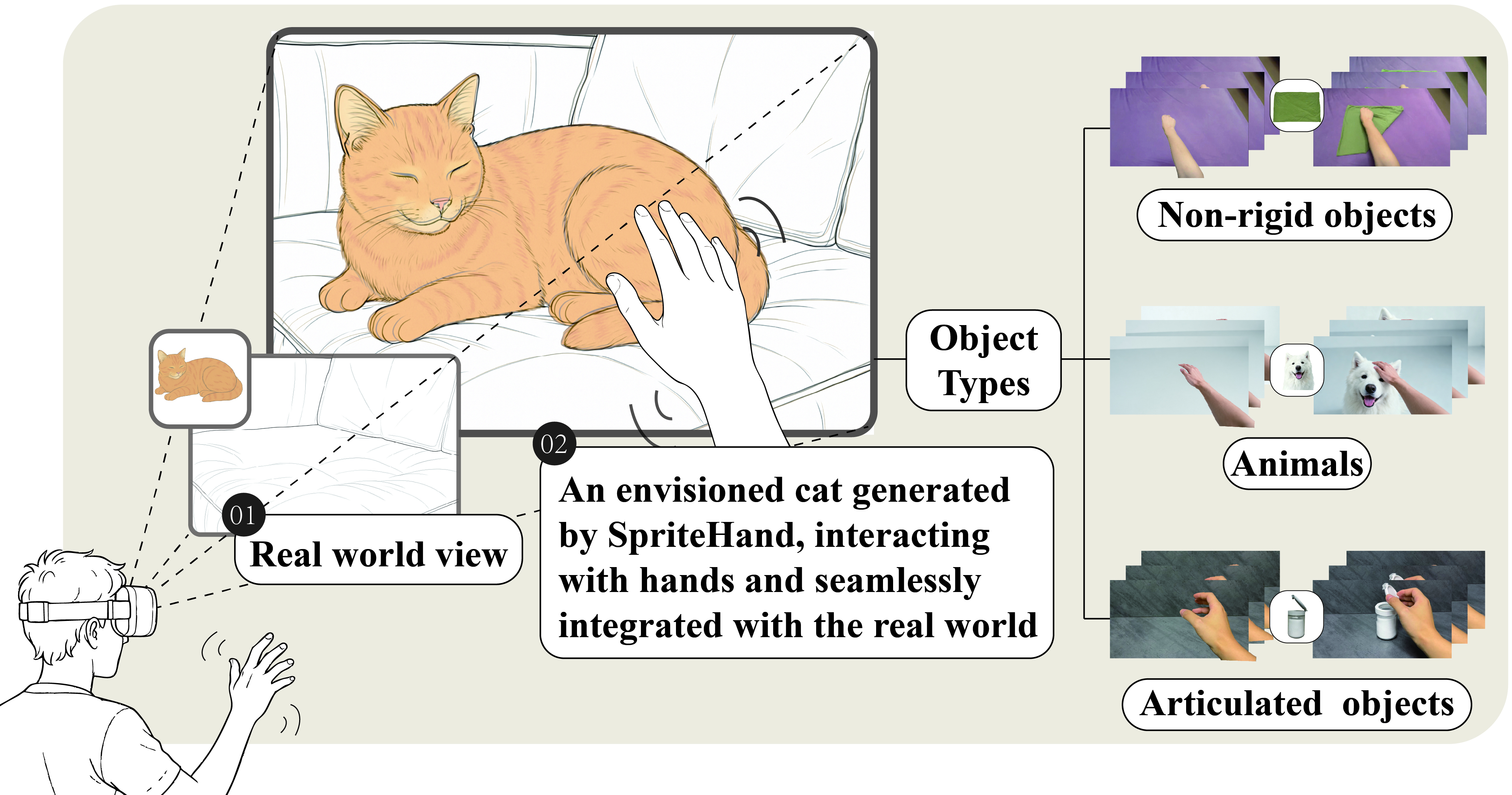}
  \caption{An envisioned application framework of SpriteHand: The user perceives the world through an egocentric camera view, either from a head-mounted display or a hand-held mobile device. SpriteHand enables various types of virtual objects to be seamlessly embedded into the live scene through a real-time generative model, dynamically adapting and responding to the user’s hand gestures within the captured environment.}
  \label{fig:teaser}
\end{figure}

\begin{abstract}
Modeling and synthesizing complex hand-object interactions remains a significant challenge, even for state-of-the-art physics engines. Conventional simulation-based approaches rely on explicitly defined rigid object models and pre-scripted hand gestures, making them inadequate for capturing dynamic interactions with non-rigid or articulated entities such as deformable fabrics, elastic materials, hinge-based structures, furry surfaces, or even living creatures. 
In this paper, we present SpriteHand, an autoregressive video generation framework for real-time synthesis of versatile hand-object interaction videos across a wide range of object types and motion patterns. SpriteHand takes as input a static object image and a video stream in which the hands are imagined to interact with the virtual object embedded in a real-world scene, and generates corresponding hand-object interaction effects in real time. Our model employs a causal inference architecture for autoregressive generation and leverages a hybrid post-training approach to enhance visual realism and temporal coherence. Our 1.3B model supports real-time streaming generation at around 18 FPS and 640×368 resolution, with an approximate 150 ms latency on a single NVIDIA RTX 5090 GPU, and more than a minute of continuous output. Experiments demonstrate superior visual quality, physical plausibility, and interaction fidelity compared to both generative and engine-based baselines.
\end{abstract}


\section{Introduction}
\label{sec:intro}

Hand-object interaction is a fundamental aspect of human experience, underpinning activities from everyday manipulation to expressive gestures. Modeling and synthesizing such interactions is crucial for a wide range of applications, including virtual reality, robotics, digital twin construction, and human-computer interaction \cite{phystwin}. However, generating realistic hand-object interactions—especially involving complex, non-rigid, or articulated objects—remains a long-standing challenge \cite{xie2024physgaussian,zhang2024physdreamer}.

Traditional simulation-based approaches rely on manually defined object models, kinematic constraints, and pre-scripted hand gestures \cite{cao2021reconstructing}. While effective for rigid and constrained scenarios, these methods struggle to capture the rich variability of real-world interactions involving deformable fabrics, elastic materials, hinge-based mechanisms, furry textures, or even living creatures \cite{10.1145/3706598.3713882}. Physics engines, despite their increasing sophistication, are often limited by their reliance on hand-tuned parameters, difficulty in modeling soft materials, and poor scalability to diverse object types and interaction contexts. 

Recent advances in generative video models have opened up new opportunities for directly learning complex interaction dynamics from large-scale real-world data \cite{akkerman2025interdyn}. Such models shift the paradigm from rule-based simulation to data-driven synthesis, enabling more flexible and visually plausible video generation. 
Recent work on accelerating video diffusion models \cite{lin2025autoregressive,yin2025causvid,hacohen2024ltx} and enabling fully streaming inference \cite{yin2025causvid} further strengthens this possibility.

In this technical context, we introduce SpriteHand, an autoregressive video generation framework designed for real-time synthesis of versatile hand-object interactions as a novel paradigm for generative interaction synthesis. SpriteHand takes as input a static object image and a video stream in which hand motions are imagined to interact with the virtual object in a real-world scene. It then produces realistic interaction effects in a streaming manner. 

To achieve this, we first train a bidirectional diffusion transformer (DiT) for hand–object interaction synthesis, and then employ a hybrid post-training strategy with self-forcing rollout \cite{huang2025selfforcing} strategy combining distribution matching \cite{yin2024onestep,yin2024improved} and adversarial training \cite{lin2025autoregressiveadversarialposttrainingrealtime}. Such a post-training process improves the video quality and physical plausibility, and reduces drift and quality degradation in long-sequence inference.


Our refined causal model built on Wan 2.1 \cite{wan2025} 1.3B achieves real-time generation at around 18 FPS and 640×368 resolution, with an average latency of around 150 ms on a single NVIDIA GTX 5090 GPU, while preserving generation quality comparable to the bidirectional teacher and substantially superior to the naive causal baseline. The system supports continuous generation for more than one minute, making it practical for interactive scenarios. 

Our paper contributes to: 1) a generative paradigm for real-time synthesis of versatile hand–object interaction, 2) a technical framework that transforms bidirectional diffusion models into high-quality causal generators via refinement, and 3) empirical validation of the feasibility of real-time generative hand–object interaction.




\section{Related Work}
\label{sec:rw}

\subsection{Hand-Object Interaction}
Reconstructing and generating hand-object interactions (HOI) from images and videos is a fundamental task in understanding human-computer interaction. This topic has garnered significant attention in computer vision communities and is now a major focus of ongoing research.

Early methods for HOI reconstruction relied on optimization over parametric models, such as combining the MANO hand model\cite{Romero2017EmbodiedH} with 3D CAD object models. However, these approaches are often computationally expensive and complex\cite{Hasson2019LearningJR, cao2021reconstructing}. Recent advances in deep learning have enabled more effective HOI reconstruction\cite{chan2022efficient,niemeyer2021giraffe}, for example by using dual-branch networks to separately predict hand and object poses, or by employing end-to-end models to directly produce joint outputs.

In parallel, HOI generation has emerged as a promising direction for synthesizing realistic interaction images or motion sequences, with broad applications in virtual humans, visual content creation, and interactive systems. Recent advances in diffusion models have significantly improved image and video synthesis capabilities. Some studies focus on Text-to-HOI generation, where diffusion models are guided by textual descriptions to produce hand-object interaction (HOI) motion sequences\cite{Christen2024DiffH2ODS,Cha2024Text2HOIT3}. However, these approaches often struggle to generate long, temporally consistent sequences and exhibit limitations in physical plausibility and realism. 

\subsection{Video Diffusion Model}
Recent advances in diffusion models have significantly improved generative capabilities in image synthesis, and these successes have been extended to the video domain. Early works such as Video Diffusion Models (VDM)\cite{Ho2022VideoDM} adapt 2D diffusion techniques by modeling temporal consistency through frame-wise conditioning or latent space propagation. Methods like Video Latent Diffusion Models (LDM) \cite{Rombach2021HighResolutionIS} project video data into a lower-dimensional latent space, enabling efficient long-range generation. Recently, video generation models based on the DiT architecture\cite{Peebles2022DiT} have further advanced the capability to generate high-quality, long-duration, and controllable videos by integrating multi-modal conditions such as text, images, and motion trajectories\cite{Yang2024CogVideoXTD, wan2025, vace}.

\subsection{Realtime Generative Video Interaction}
With the continuous advancement of video generation models, interactive generative video (IGV) has emerged as a research focus, demonstrating broad application prospects in domains such as gaming, embodied AI, and autonomous driving\cite{yu2025survey, valevski2024diffusion, che2024gamegen, zhu2024irasim, qin2024worldsimbench, hu2309gaia, lu2024wovogen}. Although diffusion-based video generation methods achieve high-quality results, they still face challenges in real-time performance; to address this, some studies have explored integrating diffusion models with autoregressive models\cite{chen2024diffusion, deng2024causal, yin2025slow}, which has to some extent improved generation efficiency.

Teacher Forcing (TF) \cite{gao2024tf} trains next-frame prediction on ground-truth contexts, leading to exposure bias during inference, while Diffusion Forcing (DF) \cite{chen2024df} alleviates this gap by introducing stochastic temporal perturbations to stabilize long rollouts.
Self Forcing (SF) \cite{huang2025sf} further aligns training with inference by conditioning each frame on the model’s own generated outputs, enabling real-time, temporally stable video generation but at the cost of reduced training efficiency.

\section{Method}
SpriteHand aims to achieve real-time video-to-video rerender, where the model receives a control video containing only hand motion and produces a corresponding interaction video in which the hand naturally manipulates a virtual object. The goal is to synthesize temporally coherent and physically plausible hand–object interactions at interactive frame rates. We achieve this by first training a bidirectional diffusion transformer (DiT) for hand–object interaction synthesis, and then distilling it into a autoregressive model that supports continuous, low-latency generation.

\begin{figure*}[t]
  \centering
  \includegraphics[width=\textwidth]{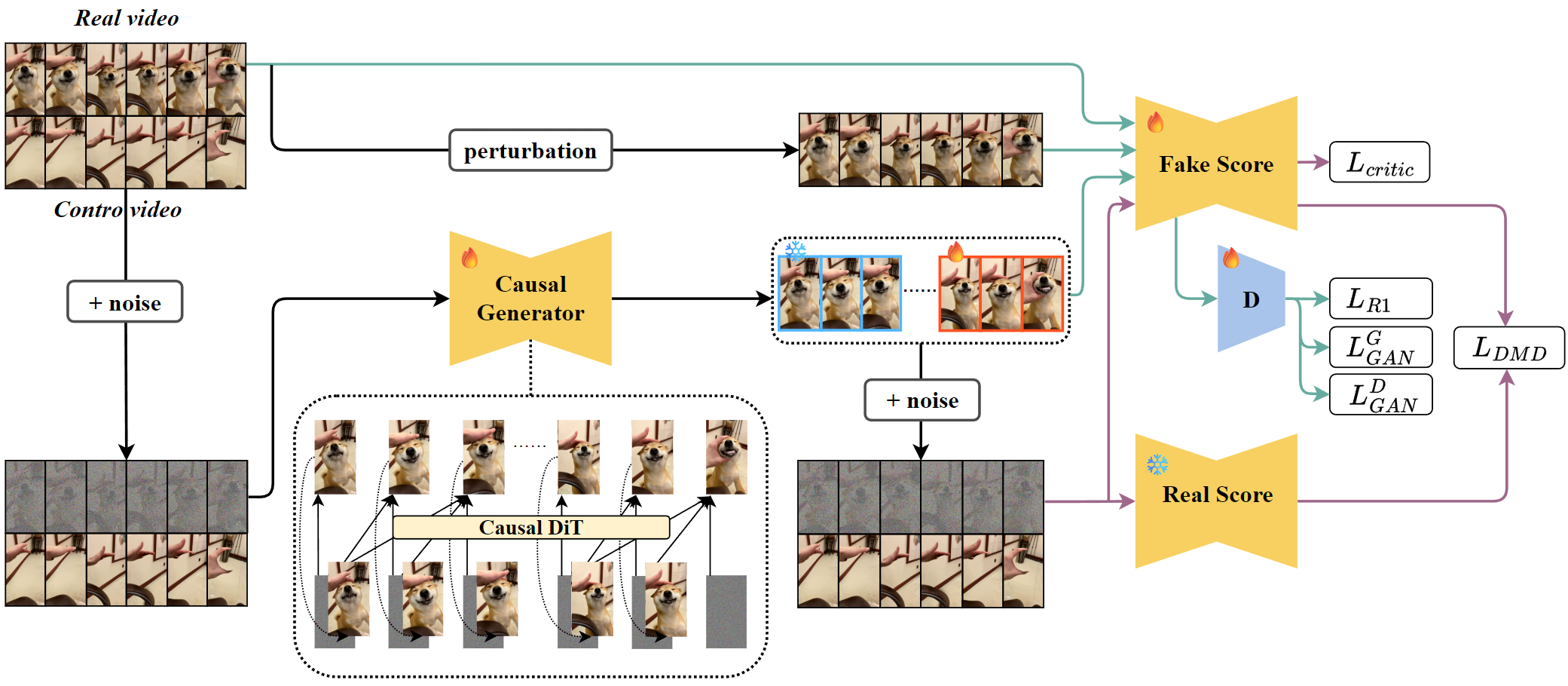}
  \caption{The overall training pipeline of SpriteHand.}
  \label{met:pipeline}
\end{figure*}

\subsection{Base Model}

We leverage Wan 2.1~\cite{wan2025wanopenadvancedlargescale}, a bidirectional video diffusion transformer trained on large-scale video datasets, as a high-quality backbone and adapt it to our hand–object interaction video-to-video generation setting. 
Our task is defined as follows: given an initial frame \(I_0\) that contains both the hand and the target object, and a control video \(C_{1:T} = \{c_1, \ldots, c_T\}\) that records a sequence of imagined hand motions without the object, the goal is to synthesize an interaction video \(V_{1:T} = \{v_1, \ldots, v_T\}\) that begins from \(I_0\) and depicts realistic, temporally coherent hand–object interactions guided by \(C_{1:T}\).

To adapt Wan 2.1 to this setting, we first encode both \(I_0\) and \(C_{1:T}\) into the latent space using the pretrained 3D variational autoencoder (VAE). The initial frame latent and the latent sequence of the control video are concatenated to the input latent in the channel dimension. We modify the first convolutional layer of the diffusion transformer to accommodate the additional input channels, initializing the newly added weights to zero to preserve the pretrained behavior. The 3D transformer is fully fine-tuned on our paired hand–object data to learn spatio-temporal correspondences between the control motion and object interaction dynamics. 

\subsection{Causal Video Generation}

While the bidirectional Wan 2.1 backbone provides high-quality global coherence, it relies on access to both past and future frames and therefore cannot support real-time streaming. 
To enable online autoregressive synthesis, we train a causal variant of the model that follows the block-causal transformer design introduced in CausVid~\cite{yin2025causvid}. 
Each latent frame constitutes a single attention block, within which tokens attend bidirectionally, while cross-frame attention is restricted by a block-causal mask so that tokens from the current frame \(t\) can only attend to tokens from frames \(1,\ldots,t\). 
Formally, the causal self-attention is defined as
\begin{equation}
    \mathrm{Attn}(Q_t, K_{\leq t}, V_{\leq t}) =
    \mathrm{Softmax}\!\left(
        \frac{Q_t K_{\leq t}^\top}{\sqrt{d}} + M_t
    \right)V_{\leq t},
\end{equation}
where \(M_t\) is the block-causal mask that prevents access to future-frame tokens.
This formulation enforces strict temporal causality while maintaining intra-frame bidirectionality.

Given a latent video sequence \(z_{1:T}\) and a time-dependent conditioning sequence \(c_{1:T}\) (the encoded reference frame and control video), the causal model factorizes the conditional generation process as:
\begin{equation}
    p_\theta(z_{1:T} \mid c_{1:T}) 
    = \prod_{t=1}^{T} p_\theta(z_t \mid z_{<t}, c_{\leq t}),
\end{equation}
, and is trained with a standard diffusion-based noise-prediction loss:
\begin{equation}
    \mathcal{L}(\theta) =
    \mathbb{E}_{z_{1:T},\,t,\,\epsilon}\!
    \left[
        \|\epsilon - \epsilon_\theta(\tilde{z}_t, t, z_{<t}, c_{\leq t})\|_2^2
    \right],
\end{equation}
, where Gaussian noise \(\epsilon\) is added to the latent inputs and the model predicts the corruption residual in a teacher-forcing manner.

During inference, cached key–value pairs from past frames are reused across attention computations, allowing each new frame to be generated in constant time by attending only to its causal history.

\subsection{Post-Training for Causal Refinement}

While the causal model enables real-time autoregressive synthesis, it inevitably suffers from the train–inference gap caused by teacher-forced training and gradual quality degradation in long-horizon video generation \cite{huang2025selfforcing}. To address this problem, we introduce a post-training stage to improve the quality of causal generation, which bridges the training-inference gap by combining self-forcing \cite{huang2025selfforcing} inference with distribution matching distillation (DMD) \cite{yin2024onestep,yin2024improved} and adversarial training \cite{lin2025diffusionadversarialposttrainingonestep,lin2025autoregressiveadversarialposttrainingrealtime}. 

\textbf{Self-Forcing Rollout.}
We unroll autoregressive video generation during post-training instead of relying on teacher-forced next-frame prediction \cite{huang2025selfforcing}. At each iteration, the causal generator $G$ starts from Gaussian noise for each frame and performs few-step diffusion denoising along a fixed timestep subsequence $\{t_1,\ldots,t_S\}$. Frames are generated sequentially using key–value caching so that each latent $z_t$ is denoised conditioned on previously self-generated frames $z_{<t}$. To maintain efficiency, gradients are propagated only through a randomly selected denoising step and through the last $k$ frames of the autoregressive rollout, while all earlier steps and frames are detached. This stochastic dual truncation scheme faithfully simulates the inference-time causal process while controlling memory and computation.

\textbf{Distribution Matching Distillation.}
To regularize the causal generator under its own rollout distribution, we employ distribution matching distillation (DMD)~\cite{yin2024onestep,yin2024improved}, which aligns two bidirectional diffusion networks: a frozen real score model $f_{\text{real}}$ and a trainable fake score model $f_{\text{fake}}$. Given self-generated latents $\hat{z}$ from the causal model $G$ using self forcing, we add Gaussian noise $\epsilon \sim \mathcal{N}(0,I)$ with scale $\sigma_t$ to obtain $\tilde{z}_t = \hat{z} + \sigma_t \epsilon$. Both networks predict denoising residuals on the same noisy latent, and the fake score is optimized to match the real score:
\begin{equation}
    \mathcal{L}_{\mathrm{DMD}} =
    \frac{1}{2}\,
    \mathbb{E}_{\hat{z}, \, t, \, \epsilon}
    \!\left[
        \|
            f_{\text{fake}}(\tilde{z}_t, t, c_{\le t})
            - f_{\text{real}}(\tilde{z}_t, t, c_{\le t})
        \|_2^2
    \right].
\end{equation}
Minimizing this objective enforces the gradient of the conditional Kullback–Leibler divergence 
$\nabla_{z_t}\mathrm{KL}(p_{\text{fake}}\|p_{\text{real}})$ 
to approach zero, thus constraining the causal generator’s induced distribution to stay aligned with the pretrained diffusion manifold.

In addition, we apply a critic loss that constrains the fake score to behave as a valid diffusion model on the generator-induced distribution. Given $\tilde{z}_t$ constructed as above, the critic predicts a residual in latent space and is trained toward the constructed flow target:
\begin{equation}
    \mathcal{L}_{\mathrm{critic}} =
    \mathbb{E}_{\hat{z},\,t,\,\epsilon}
    \!\left[
        \big\|
            f_{\text{fake}}(\tilde{z}_t, t, c_{\le t})
            - (\epsilon - \hat{z}_t)
        \big\|_2^2
    \right].
\end{equation}

This regularization maintains smooth gradient dynamics in the fake score network, preventing mode collapse and ensuring reliable gradients for both DMD alignment and subsequent adversarial refinement.

\textbf{Adversarial Refinement.}
Following the adversarial post-training paradigm~\cite{lin2025diffusionadversarialposttrainingonestep}, 
we further refine the causal generator through discriminator-guided perceptual learning. The discriminator $D$ takes intermediate layer features of the fake score network $f_{\text{fake}}$ as input,
and outputs a classification logit that discriminates between real latent $z_t$ and generated sequences $\hat{z}_t$ produced by the causal generator $G$ under self-forcing rollout.
The critic is trained using a non-saturating logistic formulation:

\begin{equation}
\begin{aligned}
    \mathcal{L}_{\mathrm{GAN}}^{D}
    &=
    \mathbb{E}_{z_t}\!\left[-\log D(z_t, c_{\le t})\right] \\
    &+
    \mathbb{E}_{\hat{z}_t}\!\left[-\log \big(1 - D(\hat{z}_t, c_{\le t})\big)\right].
\end{aligned}
\end{equation}

To stabilize adversarial optimization, we adopt the approximated R1 regularization~\cite{roth2017stabilizingtraininggenerativeadversarial,lin2025diffusionadversarialposttrainingonestep},
which enforces local smoothness of the critic around real latents through small Gaussian perturbations:
\begin{equation}
    \mathcal{L}_{\mathrm{R1}}
    =
    \mathbb{E}_{z_t}\!\left[
        \|D(z_t, c_{\le t}) - D(\mathcal{N}(z_t, \sigma I), c_{\le t})\|_2^2
    \right].
\end{equation}

The causal generator $G$ is optimized with the complementary adversarial objective that encourages generated latents to appear realistic to the critic:
\begin{equation}
    \mathcal{L}_{\mathrm{GAN}}^{G}
    =
    \mathbb{E}_{\hat{z}_t}\!\left[-\log D(\hat{z}_t, c_{\le t})\right].
\end{equation}


By jointly optimizing DMD loss, critic loss, and adversarial loss, the network ensures self-consistent gradients across diffusion and adversarial objectives, enhancing temporal coherence and perceptual realism during long-horizon causal generation.

\subsection{Training and Inference Procedure}

\textbf{Training.}
SpriteHand is trained in three sequential stages.
First, we finetune the bidirectional Wan 2.1 backbone on paired hand-object interaction data to establish a high-quality generative prior. Then we trained a causal transformer with the same model structure as Wan 2.1 under teacher forcing, enabling autoregressive video synthesis conditioned on control hand video sequences. Finally, we finetune the causal model from the previous stage with self-forcing generation, combining DMD, critic regularization, and adversarial refinement to bridge the train–inference gap for the causal generator.

In the refinement stage, as shown in Fig. \ref{met:pipeline}, the causal generator $G$ and the fake-score critic $f_{\text{fake}}$ are trained alternately, where the generator $G$ is trained once every six steps.
When training the generator, only $\mathcal{L}_{\mathrm{DMD}}$ and $\mathcal{L}_{\mathrm{GAN}}^{G}$ are optimized correspondingly, while at the critic training steps, $\mathcal{L}_{\mathrm{critic}}$, $\mathcal{L}_{\mathrm{GAN}}^{D}$, and $\mathcal{L}_{\mathrm{R1}}$ are optimized, with the generator $G$ frozen. For the self-forcing generation, each training sequence is truncated to its last 31 frames for backpropagation, and stochastic timestep sampling is used for DMD to improve temporal diversity. The weights of $\mathcal{L}_{\mathrm{DMD}}$, $\mathcal{L}_{\mathrm{critic}}$, $\mathcal{L}_{\mathrm{GAN}}^{G}$, $\mathcal{L}_{\mathrm{GAN}}^{D}$, and
$\mathcal{L}_{\mathrm{R1}}$ are set to 1.0, 1.0, 0.1, 0.05, and 100.0, respectively.

\textbf{Inference.} SpriteHand performs real-time synthesis through a streaming inference architecture that incrementally generates short temporal blocks while reusing cached computation across modules. 
The control video is divided into a bootstrap frame and successive 4-frame blocks; each block reuses cached latent features, attention states, and conditioning embeddings from previous steps. A hierarchical KV Cache \cite{yin2025causvid} is maintained across all causal attention layers, 
where new query tokens at frame $t$ attend only to previously stored $K_{\le t}$ and $V_{\le t}$ tensors on GPU memory. In addition, cached encoder–decoder states allow latent features to be propagated across blocks without recomputing intermediate representations.

We further modify the VAE to support streaming inference by decoupling temporal encoding and caching intermediate latent states. Instead of re-encoding the full sequence, the VAE processes partial clips and reuses its internal buffers for subsequent frames. Alternatively, to further accelerate decoding and strengthen temporal consistency, the standard 3D VAE can be replaced with a more compact autoencoder structure, such as the Tiny Autoencoder \cite{BoerBohan2025TAEHV}, which comprises 2D Convolutional Blocks with skip structures.

\section{Experiments}

\subsection{Datasets}

We construct a large-scale paired dataset of hand–object interaction videos to train and evaluate SpriteHand. The dataset spans three non-trivial hand-to-object interaction domains covering the interaction with non-rigid objects, articulated objects, and animals, which are hard to implement with conventional game and graphical engines.

\textbf{(1) Non-rigid objects.} 
The non-rigid object dataset comprises 954 samples of deformable and elastic materials (e.g., fabrics, rubber toys, flexible surfaces) drawn from Taste-Rob \cite{zhao2025taste-rob}. This dataset primarily captures first-person hand interactions with various fabrics and garments, such as grasping, folding, pulling, stretching, squeezing, twisting, rubbing, and flattening. Hand masks are extracted via SAM2 \cite{ravi2024sam2} and EgoHOS \cite{zhang2022egohos}, followed by attentive inpainting \cite{sun2025attentive} to restore backgrounds. 

\textbf{(2) Articulated objects.}
The articulated object dataset comprises 618 samples of hinge- and link-based structures (doors, lids, foldables) curated from Arctic \cite{fan2023arctic}. Interactions include opening, closing, rotating, flipping, pushing, pulling, and adjusting joints. Objects are obtained with SAM2 \cite{ravi2024sam2} and removed using the VACE framework \cite{jiang2025vace} built on Wan2.1-1.3B \cite{wan2025}.

\textbf{(3) Animals.}
The animal dataset comprises 491 samples of live animals interacting with first-person hands sourced from the internet, including actions such as petting, waving, feeding, and gentle pushing. Animals are obtained with SAM2 \cite{ravi2024sam2} and removed using the VACE framework \cite{jiang2025vace} built on Wan2.1-1.3B \cite{wan2025}.

\subsection{Evaluation Metrics}
We report Fréchet Inception Distance (FID) \cite{Seitzer2020FID} and Fréchet Video Distance (FVD) \cite{unterthiner2019accurategenerativemodelsvideo} together with six auxiliary metrics - aesthetic quality (AQ), Background Consistency (BC), dynamic degree (DD), imaging quality (IQ), motion smoothness (MS), and subject consistency (SC) - from VBench-I2V \cite{huang2023vbench}. FID evaluates the distributional gap between generated videos and real videos, while FVD measures temporal and spatial consistency across video sequences in terms of motion and appearance fidelity.
VBench-I2V provides a complementary perceptual evaluation of generated videos across motion quality, dynamism, aesthetic appeal, and structural fidelity.
We also report the inference latency of the first frame (only for the causal models) and the average inference speed (in FPS) under a unified resolution of (640, 368) on a single NVIDIA RTX 5090 GPU, as an indicator of the model’s real-time performance.

\begin{table*}[t]
\centering
\resizebox{0.8\textwidth}{!}{
\begin{tabular}{lcccccccccc}
\toprule
Method      &  FID↓ & FVD↓ & AQ↑ & BC↑ &DD↑ &  IQ↑  & MS↑ & SC↑ & Latency (s)↓ & FPS↑ \\
\midrule
VACE (Golden)     & 23.66 &  13.24 & 0.39 & 0.95 & 0.88 & 0.62 & 0.99 & 0.91  & / & 0.48 \\
Bidirectional     & 21.24 &  16.26 & 0.40 & 0.94 & 0.97 & 0.65 & 0.98 & 0.90  & / & 1.12 \\
Causal            & 57.11 & 175.75 & 0.35 & 0.92 & 0.88 & 0.57 & 0.98 & 0.84  & 0.21 & 11.25 \\
Causal-Refine     & 38.46 &  26.08 & 0.37 & 0.93 & 0.91 & 0.60 & 0.98 & 0.89  & 0.21 & 11.25 \\
Causal-Refine-TAE & 42.09 &  45.84 & 0.37 & 0.92 & 0.82 & 0.59 & 0.98 & 0.87  & 0.15 & 17.75 \\

\bottomrule
\end{tabular}}
\caption{Quantitative Results. FID/FVD measure visual and temporal fidelity ($\downarrow$ better). 
AQ, BC, DD, IQ, MS, and SC are VBench-I2V perceptual metrics ($\uparrow$ better). 
Latency and FPS reflect real-time responsiveness ($\downarrow$ latency, $\uparrow$ FPS).
Mean scores reported.}
\label{tab:main_comparison}
\end{table*}


\begin{table}[t]
\centering
\small
\begin{tabular}{lccccc}
\toprule
Category & Bidir & Causal & Refine & R+TAE & VACE \\
\midrule
Non-rigid   & 24.77 & 42.13  & 43.60 & 49.28 & 20.62 \\
Animal      & 47.95 & 149.90 & 72.85 & 78.38 & 65.10 \\
Articulated & 42.67 & 106.92 & 77.08 & 81.27 & 50.83 \\
\bottomrule
\end{tabular}
\caption{FID$\downarrow$ across categories and model variants. 
Bidir = Bidirectional, Refine = Causal-Refine, R+TAE = Causal-Refine-TAE.}
\label{tab:fid}
\end{table}

\begin{table}[t]
\centering
\small
\begin{tabular}{lccccc}
\toprule
Category & Bidir & Causal & Refine & R+TAE & VACE \\
\midrule
Non-rigid   & 19.32 & 85.73  & 52.18 & 100.59 & 11.04 \\
Animal      & 33.68 & 477.94 & 34.87 & 89.76  & 40.69 \\
Articulated & 26.06 & 201.91 & 30.65 & 59.28  & 9.44  \\
\bottomrule
\end{tabular}
\caption{FVD$\downarrow$ across categories and model variants.}
\label{tab:fvd}
\end{table}

\begin{table}[t]
\centering
\small
\begin{tabular}{lccccc}
\toprule
Category & Bidir & Causal & Refine & R+TAE & VACE \\
\midrule
Non-rigid   & 0.9845 & 0.9853 & 0.9840 & 0.9819 & 0.9942 \\
Animal      & 0.9653 & 0.9702 & 0.9679 & 0.9686 & 0.9821 \\
Articulated & 0.9712 & 0.9727 & 0.9709 & 0.9714 & 0.9880 \\
\bottomrule
\end{tabular}
\caption{Motion Smoothness$\uparrow$ across categories and model. 
Higher is better.}
\label{tab:smooth}
\end{table}

\subsection{Quantitative Evaluation}
We compare five model settings: the Bidirectional, Causal, Causal-Refine, Causal-Refine-TAE, and an offline “oracle’’ baseline adapted from VACE \cite{jiang2025vace}. The former three settings correspond to three stages in Sec. 3, while Causal-Refine-TAE refers to an accelerated variant with a replaced tiny autoencoder \cite{BoerBohan2025TAEHV}.
Since no prior work targets our task, VACE serves as an upper bound with privileged information (e.g., the ground-truth mask indicating the proper region for modification), while our internal variants reflect progressively stronger forms of causal adaptation.

Across metrics, we observe consistent trends (Table~\ref{tab:main_comparison}). The Bidirectional teacher achieves the best fully generative scores, while the naive Causal model drops sharply in FID/FVD due to the difficulty of autoregressive generation. Introducing our refinement stage dramatically improves temporal fidelity (e.g., FVD 175.75 → 26.08) and mitigates appearance drift, and the accelerated Causal-Refine-TAE maintains similar quality while substantially improving latency and FPS.

Compared with the oracle baseline, VACE unsurprisingly leads in FID/FVD, yet our refined variants narrow the gap considerably in perceptual metrics such as AQ, MS, and SC, and outperform it by a large margin in runtime due to full streaming compatibility. This shows that SpriteHand approaches offline fidelity while preserving real-time responsiveness.

Category-wise, non-rigid interactions are easiest, articulated cases moderately challenging, and animal interactions most difficult, with higher-frequency motion increasing FID/FVD across all models. Both refined variants consistently outperform the naive Causal baseline in every category, and all causal models maintain high motion smoothness (0.96–0.99), indicating that the main challenge lies not in jitter suppression but in long-horizon hand–object alignment.

Overall, quantitative results demonstrate that SpriteHand substantially boosts causal generation quality, approaches offline baselines, and achieves real-time performance essential for interactive mixed-reality use.

\subsection{Qualitative Analysis}
\begin{figure*}[t]
  \centering
  \includegraphics[width=\textwidth]{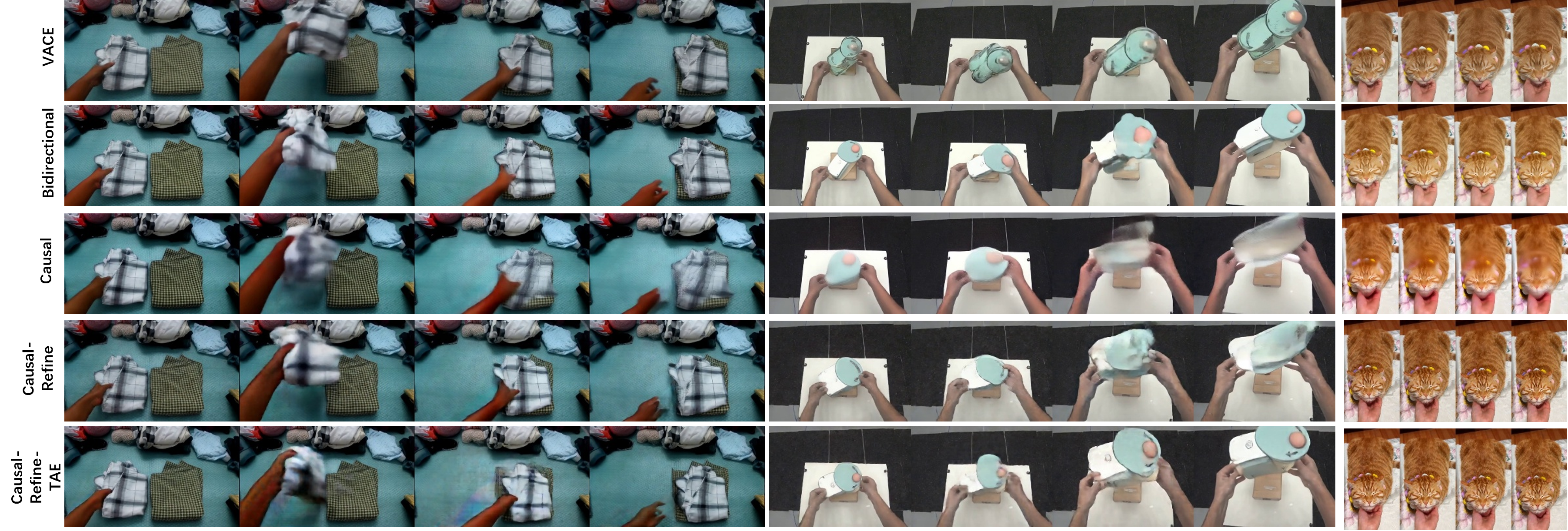}
  \caption{Qualitative comparisons of the SpriteHand framework across four model variants and the "oracle" baseline.}
  \label{fig:cross}
\end{figure*}

Fig. \ref{fig:cross} provides a qualitative overview of the proposed SpriteHand framework. When comparing the five model variants across the three interaction categories, we observe that VACE and the Bidirectional teacher produce the highest visual fidelity, yet rely on offline processing and therefore cannot support real-time interaction. In contrast, the naive Causal model, while fully streaming-compatible, suffers from drift, softened textures, and weakened temporal consistency. The refined variants — Causal-Refine and Causal-Refine-TAE — significantly narrow this quality gap: object geometry, hand–object alignment, and temporal coherence appear close to the Bidirectional teacher while still maintaining real-time responsiveness.

\begin{figure}[t]
  \centering
  \includegraphics[width=\linewidth]{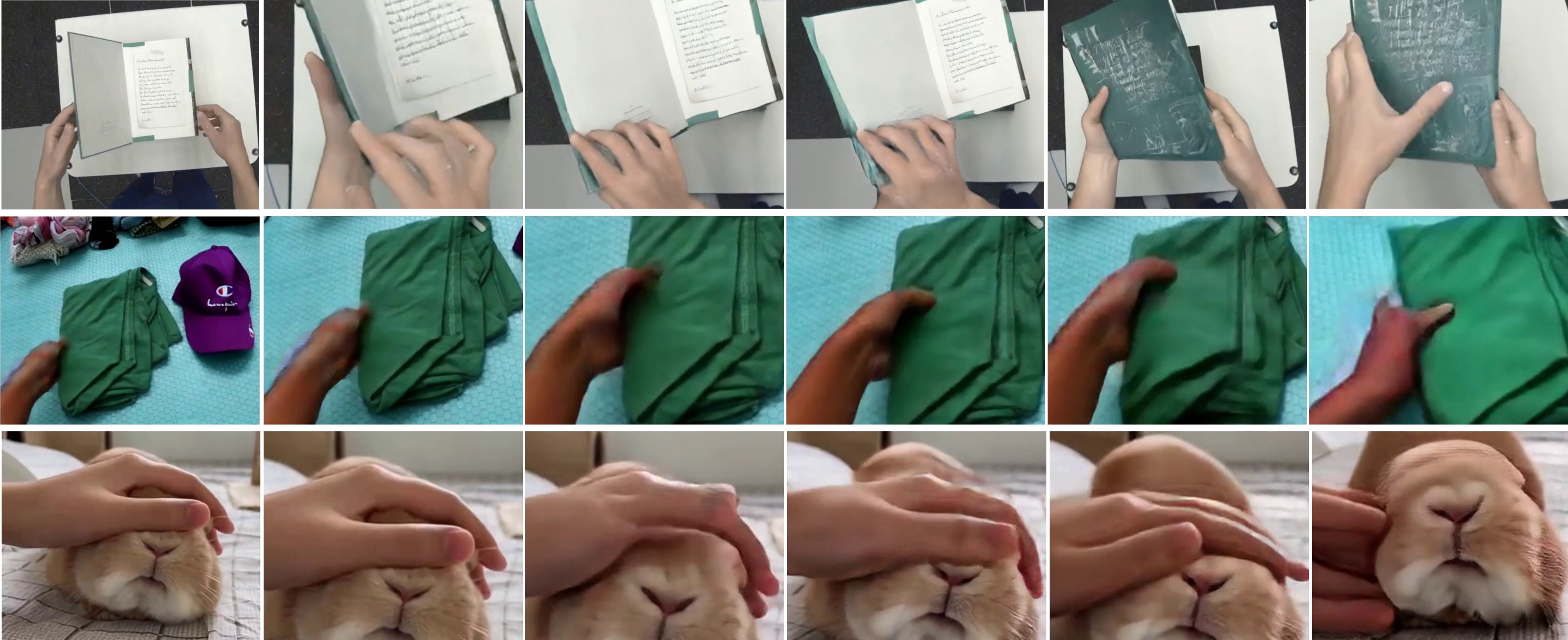}
  \caption{Interaction details cross three HOI domains generated by Causal-Refine-TAE.}
  \label{fig:details}
\end{figure}

Moreover, we observed that both Causal-Refine and its accelerated TAE counterpart produce physically plausible spatial details and interaction dynamics than the naive causal baseline (Fig.~\ref{fig:details}), demonstrating that SpriteHand successfully restores fine-grained gesture–object coupling under streaming inference.

\begin{figure*}[t]
  \centering
  \includegraphics[width=\textwidth]{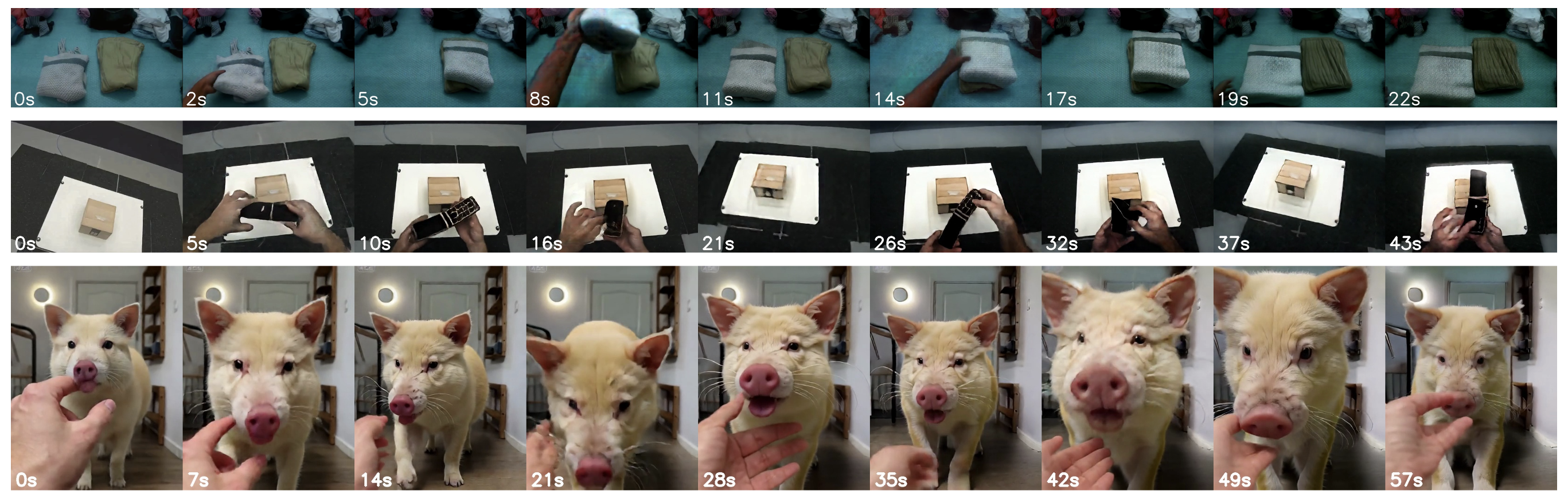}
  \caption{Qualitative results of long video samples. We synthesize long video samples by concatenating particular video samples forward and backward alternatively 5 times, which exceeds the maximum length in training. Results are produced by the Causal-Refine model, showing good generalizability of our approach in long-horizon inference. }
  \label{fig:long_sample}
\end{figure*}

\begin{figure*}[t]
  \centering
  \includegraphics[width=\textwidth]{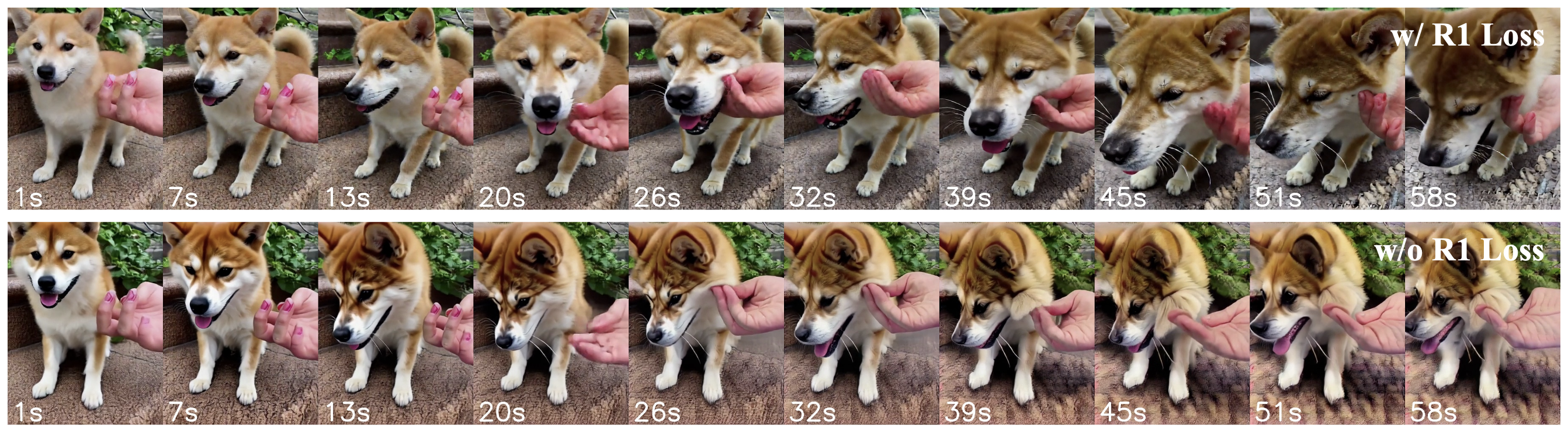}
  \caption{Ablation comparison on the use of R1 loss. Introducing R1 loss significantly improves the quality and naturalness of generated samples, and reduces the drifting and quality-degrading problem in long-sequence inference.}
  \label{fig:r1_compare}
\end{figure*}

The model exhibits strong robustness on sequences far beyond the training horizon (Fig.~\ref{fig:long_sample}). When looping video clips forward and backward to form minute-long sequences, the Causal-Refine model maintains stable appearance and consistent interaction without accumulating drift, indicating reliable long-horizon behavior crucial for interactive mixed-reality applications. Ablation study (Fig.~\ref{fig:r1_compare}) further indicates that long-horizon stability is strongly supported by the inclusion of the R1 loss: without it, extended rollout tends to accumulate texture degradation and temporal drift, whereas adding R1 substantially suppresses these artifacts and plays a key role in maintaining coherent long-sequence generation.


While overall performance is strong, we observe occasional failure cases: complex mechanical objects may be misinterpreted without prior structural cues, heavy occlusions or stacked objects can lead to incorrect topological merging, and rare animal poses may cause transient shape collapse. These cases highlight remaining challenges in modeling diverse and highly dynamic hand–object interactions.

\subsection{Human Evaluation}
To understand how users perceive the realism and responsiveness of our generative interaction framework, which synthesizes virtual object behavior directly within real video streams and operates in real time, we further conduct a human evaluation focused on perceptual factors most relevant to mixed-reality interaction.

\paragraph{Stimuli.}
We sample 15 videos generated by our causal-refined model across three hand-to-object interaction domains. For comparison, we collect 15 clips from two representative external sources: (1) recordings from commercial mixed-reality applications where a user interacts with a virtual pet, and (2) videos from prior work in which hand–object interactions are implemented via 3D reconstructed assets and scripted behaviors in a virtual engine \cite{10.1145/3706598.3713882}. All clips are standardized to 4–6 seconds, identical resolution, and presented without audio or interface elements. Participants view all 30 clips in randomized order.

\paragraph{Evaluation Criteria.}
After each clip, participants provide 5-point Mean Opinion Scores (MOS) along four perceptual dimensions: Mixed-Reality Realism, which measures how natural the interaction appears within the real scene; Interaction Plausibility, which captures the physical believability of contact and object response; Control Responsiveness, which reflects the perceived immediacy of the object's reaction to hand motion; and Reality Integration, which assesses how seamlessly the virtual content blends into the captured environment. These dimensions directly reflect our goal of achieving coherent, real-time interaction that blurs the boundary between virtual and physical elements.

\paragraph{Results.}
SpriteHand achieves competitive or higher MOS than both comparison sources (Table~\ref{tab:mos_results}). Participants report more consistent contact timing and smoother alignment between hand motion and object reaction, resulting in higher Interaction Plausibility. Control Responsiveness is also rated favorably, indicating that our causal streaming design reaches a perceptual latency comparable to real-time interactive systems. Reality Integration also favors SpriteHand, with participants noting that the generated objects appear more naturally embedded in the real scene, although the improvement is more moderate compared to other dimensions. These findings support our central claim that SpriteHand enables a new interaction paradigm in which virtual behaviors are synthesized as if they were intrinsic to the physical environment, effectively narrowing the divide between virtual and real interactions.

\begin{table}[t]
\centering
\small
\resizebox{\linewidth}{!}{
\begin{tabular}{lcccc}
\toprule
Method & MR & IP & CR & RI \\
\midrule
SpriteHand   & 3.8±1.3 & 3.8±1.2 & 4.1±1.0 & 3.9±1.2 \\
Other Sources & 2.3±1.1 & 2.5±1.2 & 2.9±1.1 & 2.4±1.2 \\
\bottomrule
\end{tabular}}
\caption{
Mean Opinion Scores (MOS; $1$--$5$ scale) across the four perceptual 
dimensions: Mixed-Reality Realism (MR), Interaction Plausibility (IP), 
Control Responsiveness (CR), and Reality Integration (RI).
}
\label{tab:mos_results}
\end{table}

\section{Conclusion}
In this work, we envisioned SpriteHand as an early step toward a new interaction paradigm in which virtual-physical interactions are generated as if they belong to the real visual world. By transforming a bidirectional diffusion transformer into a real-time autoregressive generator using self-forcing rollout, distribution matching, and adversarial post-training, SpriteHand produces physically coherent interactions from only a static object image and a streaming hand-motion video. Our experiments show that the refined causal model approaches the quality of its bidirectional teacher while maintaining low-latency, minute-long continuous generation, pointing toward a data-driven alternative to traditional simulation pipelines. We hope this work encourages further exploration of real-time generative interaction as a foundation for richer, more seamless virtual–physical experiences.

    \bibliographystyle{ieeenat_fullname}
    \bibliography{main}

@String(CVPR= {IEEE Conf. Comput. Vis. Pattern Recog.})

@String(ICCV= {Int. Conf. Comput. Vis.})

@String(AAAI = {AAAI})

@String(CVPR  = {CVPR})

@String(ICCV  = {ICCV})

@inproceedings{zhao2025taste-rob,
  title={TASTE-Rob: Advancing video generation of task-oriented hand-object interaction for generalizable robotic manipulation},
  author={Zhao, Hongxiang and Liu, Xingchen and Xu, Mutian and Hao, Yiming and Chen, Weikai and Han, Xiaoguang},
  booktitle={Proceedings of the Computer Vision and Pattern Recognition Conference},
  pages={27683--27693},
  year={2025}
}

@inproceedings{fan2023arctic,
  title={ARCTIC: A dataset for dexterous bimanual hand-object manipulation},
  author={Fan, Zicong and Taheri, Omid and Tzionas, Dimitrios and Kocabas, Muhammed and Kaufmann, Manuel and Black, Michael J and Hilliges, Otmar},
  booktitle={Proceedings of the IEEE/CVF conference on computer vision and pattern recognition},
  pages={12943--12954},
  year={2023}
}

@inproceedings{zhang2022egohos,
  title={F},
  author={Zhang, Lingzhi and Zhou, Shenghao and Stent, Simon and Shi, Jianbo},
  booktitle={European Conference on Computer Vision},
  pages={127--145},
  year={2022},
  organization={Springer}
}

@article{ravi2024sam2,
  title={SAM 2: Segment Anything in Images and Videos},
  author={Ravi, Nikhila and Gabeur, Valentin and Hu, Yuan-Ting and Hu, Ronghang and Ryali, Chaitanya and Ma, Tengyu and Khedr, Haitham and R{\"a}dle, Roman and Rolland, Chloe and Gustafson, Laura and Mintun, Eric and Pan, Junting and Alwala, Kalyan Vasudev and Carion, Nicolas and Wu, Chao-Yuan and Girshick, Ross and Doll{\'a}r, Piotr and Feichtenhofer, Christoph},
  journal={arXiv preprint arXiv:2408.00714},
  url={https://arxiv.org/abs/2408.00714},
  year={2024}
}

@article{jiang2025vace,
  title={Vace: All-in-one video creation and editing},
  author={Jiang, Zeyinzi and Han, Zhen and Mao, Chaojie and Zhang, Jingfeng and Pan, Yulin and Liu, Yu},
  journal={arXiv preprint arXiv:2503.07598},
  year={2025}
}

@article{wan2025,
      title={Wan: Open and Advanced Large-Scale Video Generative Models}, 
      author={Team Wan and Ang Wang and Baole Ai and Bin Wen and Chaojie Mao and Chen-Wei Xie and Di Chen and Feiwu Yu and Haiming Zhao and Jianxiao Yang and Jianyuan Zeng and Jiayu Wang and Jingfeng Zhang and Jingren Zhou and Jinkai Wang and Jixuan Chen and Kai Zhu and Kang Zhao and Keyu Yan and Lianghua Huang and Mengyang Feng and Ningyi Zhang and Pandeng Li and Pingyu Wu and Ruihang Chu and Ruili Feng and Shiwei Zhang and Siyang Sun and Tao Fang and Tianxing Wang and Tianyi Gui and Tingyu Weng and Tong Shen and Wei Lin and Wei Wang and Wei Wang and Wenmeng Zhou and Wente Wang and Wenting Shen and Wenyuan Yu and Xianzhong Shi and Xiaoming Huang and Xin Xu and Yan Kou and Yangyu Lv and Yifei Li and Yijing Liu and Yiming Wang and Yingya Zhang and Yitong Huang and Yong Li and You Wu and Yu Liu and Yulin Pan and Yun Zheng and Yuntao Hong and Yupeng Shi and Yutong Feng and Zeyinzi Jiang and Zhen Han and Zhi-Fan Wu and Ziyu Liu},
      journal = {arXiv preprint arXiv:2503.20314},
      year={2025}
}

@inproceedings{cao2021reconstructing,
  title={Reconstructing hand-object interactions in the wild},
  author={Cao, Zhe and Radosavovic, Ilija and Kanazawa, Angjoo and Malik, Jitendra},
  booktitle={Proceedings of the IEEE/CVF international conference on computer vision},
  pages={12417--12426},
  year={2021}
}

@article{Hasson2019LearningJR,
  title={Learning Joint Reconstruction of Hands and Manipulated Objects},
  author={Yana Hasson and G{\"u}l Varol and Dimitrios Tzionas and Igor Kalevatykh and Michael J. Black and Ivan Laptev and Cordelia Schmid},
  journal={2019 IEEE/CVF Conference on Computer Vision and Pattern Recognition (CVPR)},
  year={2019},
  pages={11799-11808},
  url={https://api.semanticscholar.org/CorpusID:106404030}
}

@article{Ho2022VideoDM,
  title={Video Diffusion Models},
  author={Jonathan Ho and Tim Salimans and Alexey Gritsenko and William Chan and Mohammad Norouzi and David J. Fleet},
  journal={ArXiv},
  year={2022},
  volume={abs/2204.03458},
  url={https://api.semanticscholar.org/CorpusID:248006185}
}

@article{Rombach2021HighResolutionIS,
  title={High-Resolution Image Synthesis with Latent Diffusion Models},
  author={Robin Rombach and A. Blattmann and Dominik Lorenz and Patrick Esser and Bj{\"o}rn Ommer},
  journal={2022 IEEE/CVF Conference on Computer Vision and Pattern Recognition (CVPR)},
  year={2021},
  pages={10674-10685},
  url={https://api.semanticscholar.org/CorpusID:245335280}
}

@article{Peebles2022DiT,
  title={Scalable Diffusion Models with Transformers},
  author={William Peebles and Saining Xie},
  year={2022},
  journal={arXiv preprint arXiv:2212.09748},
}

@article{Yang2024CogVideoXTD,
  title={CogVideoX: Text-to-Video Diffusion Models with An Expert Transformer},
  author={Zhuoyi Yang and Jiayan Teng and Wendi Zheng and Ming Ding and Shiyu Huang and Jiazheng Xu and Yuanming Yang and Wenyi Hong and Xiaohan Zhang and Guanyu Feng and Da Yin and Xiaotao Gu and Yuxuan Zhang and Weihan Wang and Yean Cheng and Ting Liu and Bin Xu and Yuxiao Dong and Jie Tang},
  journal={ArXiv},
  year={2024},
  volume={abs/2408.06072},
  url={https://api.semanticscholar.org/CorpusID:271855655}
}

@inproceedings{Romero2017EmbodiedH,
  title={Embodied Hands : Modeling and Capturing Hands and Bodies Together * * Supplementary Material * *},
  author={Javier Romero and Dimitrios Tzionas},
  year={2017},
  url={https://api.semanticscholar.org/CorpusID:1475882}
}

@article{vace,
    title = {VACE: All-in-One Video Creation and Editing},
    author = {Jiang, Zeyinzi and Han, Zhen and Mao, Chaojie and Zhang, Jingfeng and Pan, Yulin and Liu, Yu},
    journal = {arXiv preprint arXiv:2503.07598},
    year = {2025}
}

@article{Christen2024DiffH2ODS,
  title={DiffH2O: Diffusion-Based Synthesis of Hand-Object Interactions from Textual Descriptions},
  author={Sammy Joe Christen and Shreyas Hampali and Fadime Sener and Edoardo Remelli and Tom{\'a}s Hodan and Eric Sauser and Shugao Ma and Bugra Tekin},
  journal={SIGGRAPH Asia 2024 Conference Papers},
  year={2024},
  url={https://api.semanticscholar.org/CorpusID:268691880}
}

@article{Cha2024Text2HOIT3,
  title={Text2HOI: Text-Guided 3D Motion Generation for Hand-Object Interaction},
  author={Junuk Cha and Jihyeon Kim and Jae Shin Yoon and Seungryul Baek},
  journal={2024 IEEE/CVF Conference on Computer Vision and Pattern Recognition (CVPR)},
  year={2024},
  pages={1577-1585},
  url={https://api.semanticscholar.org/CorpusID:268819822}
}

@inproceedings{chan2022efficient,
  title={Efficient geometry-aware 3d generative adversarial networks},
  author={Chan, Eric R and Lin, Connor Z and Chan, Matthew A and Nagano, Koki and Pan, Boxiao and De Mello, Shalini and Gallo, Orazio and Guibas, Leonidas J and Tremblay, Jonathan and Khamis, Sameh and others},
  booktitle={Proceedings of the IEEE/CVF conference on computer vision and pattern recognition},
  pages={16123--16133},
  year={2022}
}

@inproceedings{niemeyer2021giraffe,
  title={Giraffe: Representing scenes as compositional generative neural feature fields},
  author={Niemeyer, Michael and Geiger, Andreas},
  booktitle={Proceedings of the IEEE/CVF conference on computer vision and pattern recognition},
  pages={11453--11464},
  year={2021}
}

@article{yu2025survey,
  title={A survey of interactive generative video},
  author={Yu, Jiwen and Qin, Yiran and Che, Haoxuan and Liu, Quande and Wang, Xintao and Wan, Pengfei and Zhang, Di and Gai, Kun and Chen, Hao and Liu, Xihui},
  journal={arXiv preprint arXiv:2504.21853},
  year={2025}
}

@article{valevski2024diffusion,
  title={Diffusion models are real-time game engines},
  author={Valevski, Dani and Leviathan, Yaniv and Arar, Moab and Fruchter, Shlomi},
  journal={arXiv preprint arXiv:2408.14837},
  year={2024}
}

@article{che2024gamegen,
  title={Gamegen-x: Interactive open-world game video generation},
  author={Che, Haoxuan and He, Xuanhua and Liu, Quande and Jin, Cheng and Chen, Hao},
  journal={arXiv preprint arXiv:2411.00769},
  year={2024}
}

@article{zhu2024irasim,
  title={Irasim: Learning interactive real-robot action simulators},
  author={Zhu, Fangqi and Wu, Hongtao and Guo, Song and Liu, Yuxiao and Cheang, Chilam and Kong, Tao},
  journal={arXiv preprint arXiv:2406.14540},
  year={2024}
}

@article{qin2024worldsimbench,
  title={Worldsimbench: Towards video generation models as world simulators},
  author={Qin, Yiran and Shi, Zhelun and Yu, Jiwen and Wang, Xijun and Zhou, Enshen and Li, Lijun and Yin, Zhenfei and Liu, Xihui and Sheng, Lu and Shao, Jing and others},
  journal={arXiv preprint arXiv:2410.18072},
  year={2024}
}

@article{hu2309gaia,
  title={GAIA-1: a generative world model for autonomous driving (2023)},
  author={Hu, Anthony and Russell, Lloyd and Yeo, Hudson and Murez, Zak and Fedoseev, George and Kendall, Alex and Shotton, Jamie and Corrado, Gianluca},
  journal={arXiv preprint arXiv:2309.17080}
}

@inproceedings{lu2024wovogen,
  title={Wovogen: World volume-aware diffusion for controllable multi-camera driving scene generation},
  author={Lu, Jiachen and Huang, Ze and Yang, Zeyu and Zhang, Jiahui and Zhang, Li},
  booktitle={European Conference on Computer Vision},
  pages={329--345},
  year={2024},
  organization={Springer}
}

@article{chen2024diffusion,
  title={Diffusion forcing: Next-token prediction meets full-sequence diffusion},
  author={Chen, Boyuan and Mart{\'\i} Mons{\'o}, Diego and Du, Yilun and Simchowitz, Max and Tedrake, Russ and Sitzmann, Vincent},
  journal={Advances in Neural Information Processing Systems},
  volume={37},
  pages={24081--24125},
  year={2024}
}

@article{deng2024causal,
  title={Causal diffusion transformers for generative modeling},
  author={Deng, Chaorui and Zhu, Deyao and Li, Kunchang and Guang, Shi and Fan, Haoqi},
  journal={arXiv preprint arXiv:2412.12095},
  year={2024}
}

@inproceedings{yin2025slow,
  title={From slow bidirectional to fast autoregressive video diffusion models},
  author={Yin, Tianwei and Zhang, Qiang and Zhang, Richard and Freeman, William T and Durand, Fredo and Shechtman, Eli and Huang, Xun},
  booktitle={Proceedings of the Computer Vision and Pattern Recognition Conference},
  pages={22963--22974},
  year={2025}
}

@inproceedings{sun2025attentive,
  title={Attentive eraser: Unleashing diffusion model’s object removal potential via self-attention redirection guidance},
  author={Sun, Wenhao and Dong, Xue-Mei and Cui, Benlei and Tang, Jingqun},
  booktitle={Proceedings of the AAAI Conference on Artificial Intelligence},
  volume={39},
  number={19},
  pages={20734--20742},
  year={2025}
}

@misc{wan2025wanopenadvancedlargescale,
      title={Wan: Open and Advanced Large-Scale Video Generative Models}, 
      author={Team Wan and Ang Wang and Baole Ai and Bin Wen and Chaojie Mao and Chen-Wei Xie and Di Chen and Feiwu Yu and Haiming Zhao and Jianxiao Yang and Jianyuan Zeng and Jiayu Wang and Jingfeng Zhang and Jingren Zhou and Jinkai Wang and Jixuan Chen and Kai Zhu and Kang Zhao and Keyu Yan and Lianghua Huang and Mengyang Feng and Ningyi Zhang and Pandeng Li and Pingyu Wu and Ruihang Chu and Ruili Feng and Shiwei Zhang and Siyang Sun and Tao Fang and Tianxing Wang and Tianyi Gui and Tingyu Weng and Tong Shen and Wei Lin and Wei Wang and Wei Wang and Wenmeng Zhou and Wente Wang and Wenting Shen and Wenyuan Yu and Xianzhong Shi and Xiaoming Huang and Xin Xu and Yan Kou and Yangyu Lv and Yifei Li and Yijing Liu and Yiming Wang and Yingya Zhang and Yitong Huang and Yong Li and You Wu and Yu Liu and Yulin Pan and Yun Zheng and Yuntao Hong and Yupeng Shi and Yutong Feng and Zeyinzi Jiang and Zhen Han and Zhi-Fan Wu and Ziyu Liu},
      year={2025},
      eprint={2503.20314},
      archivePrefix={arXiv},
      primaryClass={cs.CV},
      url={https://arxiv.org/abs/2503.20314}, 
}

@inproceedings{yin2025causvid,
    title={From Slow Bidirectional to Fast Autoregressive Video Diffusion Models},
    author={Yin, Tianwei and Zhang, Qiang and Zhang, Richard and Freeman, William T and Durand, Fredo and Shechtman, Eli and Huang, Xun},
    booktitle={CVPR},
    year={2025}
}

@inproceedings{yin2024improved,
    title={Improved Distribution Matching Distillation for Fast Image Synthesis},
    author={Yin, Tianwei and Gharbi, Micha{\"e}l and Park, Taesung and Zhang, Richard and Shechtman, Eli and Durand, Fredo and Freeman, William T},
    booktitle={NeurIPS},
    year={2024}
}

@inproceedings{yin2024onestep,
    title={One-step Diffusion with Distribution Matching Distillation},
    author={Yin, Tianwei and Gharbi, Micha{\"e}l and Zhang, Richard and Shechtman, Eli and Durand, Fr{\'e}do and Freeman, William T and Park, Taesung},
    booktitle={CVPR},
    year={2024}
}

@misc{lin2025diffusionadversarialposttrainingonestep,
      title={Diffusion Adversarial Post-Training for One-Step Video Generation}, 
      author={Shanchuan Lin and Xin Xia and Yuxi Ren and Ceyuan Yang and Xuefeng Xiao and Lu Jiang},
      year={2025},
      eprint={2501.08316},
      archivePrefix={arXiv},
      primaryClass={cs.CV},
      url={https://arxiv.org/abs/2501.08316}, 
}

@misc{lin2025autoregressiveadversarialposttrainingrealtime,
      title={Autoregressive Adversarial Post-Training for Real-Time Interactive Video Generation}, 
      author={Shanchuan Lin and Ceyuan Yang and Hao He and Jianwen Jiang and Yuxi Ren and Xin Xia and Yang Zhao and Xuefeng Xiao and Lu Jiang},
      year={2025},
      eprint={2506.09350},
      archivePrefix={arXiv},
      primaryClass={cs.CV},
      url={https://arxiv.org/abs/2506.09350}, 
}

@misc{roth2017stabilizingtraininggenerativeadversarial,
      title={Stabilizing Training of Generative Adversarial Networks through Regularization}, 
      author={Kevin Roth and Aurelien Lucchi and Sebastian Nowozin and Thomas Hofmann},
      year={2017},
      eprint={1705.09367},
      archivePrefix={arXiv},
      primaryClass={cs.LG},
      url={https://arxiv.org/abs/1705.09367}, 
}

@article{huang2025selfforcing,
  title={Self Forcing: Bridging the Train-Test Gap in Autoregressive Video Diffusion},
  author={Huang, Xun and Li, Zhengqi and He, Guande and Zhou, Mingyuan and Shechtman, Eli},
  journal={arXiv preprint arXiv:2506.08009},
  year={2025}
}

@misc {BoerBohan2025TAEHV,
  author = {Boer Bohan, Ollin},
  title = {TAEHV: Tiny AutoEncoder for Hunyuan Video},
  year = {2025},
  howpublished = {\url{https://github.com/madebyollin/taehv}},
}

@misc{Seitzer2020FID,
  author={Maximilian Seitzer},
  title={{pytorch-fid: FID Score for PyTorch}},
  month={August},
  year={2020},
  note={Version 0.3.0},
  howpublished={\url{https://github.com/mseitzer/pytorch-fid}},
}

@misc{unterthiner2019accurategenerativemodelsvideo,
      title={Towards Accurate Generative Models of Video: A New Metric \& Challenges}, 
      author={Thomas Unterthiner and Sjoerd van Steenkiste and Karol Kurach and Raphael Marinier and Marcin Michalski and Sylvain Gelly},
      year={2019},
      eprint={1812.01717},
      archivePrefix={arXiv},
      primaryClass={cs.CV},
      url={https://arxiv.org/abs/1812.01717}, 
}

@InProceedings{huang2023vbench,
     title={{VBench}: Comprehensive Benchmark Suite for Video Generative Models},
     author={Huang, Ziqi and He, Yinan and Yu, Jiashuo and Zhang, Fan and Si, Chenyang and Jiang, Yuming and Zhang, Yuanhan and Wu, Tianxing and Jin, Qingyang and Chanpaisit, Nattapol and Wang, Yaohui and Chen, Xinyuan and Wang, Limin and Lin, Dahua and Qiao, Yu and Liu, Ziwei},
     booktitle={Proceedings of the IEEE/CVF Conference on Computer Vision and Pattern Recognition},
     year={2024}
 }

@article{gao2024tf,
  title={Ca2-vdm: Efficient autoregressive video diffusion model with causal generation and cache sharing},
  author={Gao, Kaifeng and Shi, Jiaxin and Zhang, Hanwang and Wang, Chunping and Xiao, Jun and Chen, Long},
  journal={arXiv preprint arXiv:2411.16375},
  year={2024}
}

@article{chen2024df,
  title={Diffusion forcing: Next-token prediction meets full-sequence diffusion},
  author={Chen, Boyuan and Mart{\'\i} Mons{\'o}, Diego and Du, Yilun and Simchowitz, Max and Tedrake, Russ and Sitzmann, Vincent},
  journal={Advances in Neural Information Processing Systems},
  volume={37},
  pages={24081--24125},
  year={2024}
}

@article{huang2025sf,
  title={Self Forcing: Bridging the Train-Test Gap in Autoregressive Video Diffusion},
  author={Huang, Xun and Li, Zhengqi and He, Guande and Zhou, Mingyuan and Shechtman, Eli},
  journal={arXiv preprint arXiv:2506.08009},
  year={2025}
}

@inproceedings{10.1145/3706598.3713882,
author = {Li, Zisu and Li, Jiawei and Xiong, Zeyu and Zhang, Shumeng and Faruqi, Faraz and Mueller, Stefanie and Liang, Chen and Ma, Xiaojuan and Fan, Mingming},
title = {InteRecon: Towards Reconstructing Interactivity of Personal Memorable Items in Mixed Reality},
year = {2025},
isbn = {9798400713941},
publisher = {Association for Computing Machinery},
address = {New York, NY, USA},
url = {https://doi.org/10.1145/3706598.3713882},
doi = {10.1145/3706598.3713882},
booktitle = {Proceedings of the 2025 CHI Conference on Human Factors in Computing Systems},
articleno = {833},
numpages = {19},
location = {
},
series = {CHI '25}
}

@article{phystwin,
    title={PhysTwin: Physics-Informed Reconstruction and Simulation of Deformable Objects from Videos},
    author={Jiang, Hanxiao and Hsu, Hao-Yu and Zhang, Kaifeng and Yu, Hsin-Ni and Wang, Shenlong and Li, Yunzhu},
    journal={ICCV},
    year={2025}
}

@inproceedings{xie2024physgaussian,
  title={Physgaussian: Physics-integrated 3d gaussians for generative dynamics},
  author={Xie, Tianyi and Zong, Zeshun and Qiu, Yuxing and Li, Xuan and Feng, Yutao and Yang, Yin and Jiang, Chenfanfu},
  booktitle={Proceedings of the IEEE/CVF Conference on Computer Vision and Pattern Recognition},
  pages={4389--4398},
  year={2024}
}

@inproceedings{zhang2024physdreamer,
  title={Physdreamer: Physics-based interaction with 3d objects via video generation},
  author={Zhang, Tianyuan and Yu, Hong-Xing and Wu, Rundi and Feng, Brandon Y and Zheng, Changxi and Snavely, Noah and Wu, Jiajun and Freeman, William T},
  booktitle={European Conference on Computer Vision},
  pages={388--406},
  year={2024},
  organization={Springer}
}

@inproceedings{akkerman2025interdyn,
  title={InterDyn: Controllable interactive dynamics with video diffusion models},
  author={Akkerman, Rick and Feng, Haiwen and Black, Michael J and Tzionas, Dimitrios and Abrevaya, Victoria Fern{\'a}ndez},
  booktitle={Proceedings of the Computer Vision and Pattern Recognition Conference},
  pages={12467--12479},
  year={2025}
}

@article{lin2025autoregressive,
  title={Autoregressive Adversarial Post-Training for Real-Time Interactive Video Generation},
  author={Lin, Shanchuan and Yang, Ceyuan and He, Hao and Jiang, Jianwen and Ren, Yuxi and Xia, Xin and Zhao, Yang and Xiao, Xuefeng and Jiang, Lu},
  journal={arXiv preprint arXiv:2506.09350},
  year={2025}
}

@article{hacohen2024ltx,
  title={Ltx-video: Realtime video latent diffusion},
  author={HaCohen, Yoav and Chiprut, Nisan and Brazowski, Benny and Shalem, Daniel and Moshe, Dudu and Richardson, Eitan and Levin, Eran and Shiran, Guy and Zabari, Nir and Gordon, Ori and others},
  journal={arXiv preprint arXiv:2501.00103},
  year={2024}
}


\end{document}